\documentclass[runningheads]{llncs}
\usepackage[T1]{fontenc}
\usepackage{graphicx}
\usepackage{amsmath,amsfonts}
\usepackage{algorithmic}
\usepackage{algorithm}
\usepackage{array}
\usepackage{textcomp}
\usepackage{stfloats}
\usepackage{url}
\usepackage{verbatim}
\usepackage{graphicx}
\usepackage{cite}
\usepackage[table]{xcolor}
\usepackage{footnote}

\newcommand{\astfootnote}[1]{%
\let\oldthefootnote=\thefootnote%
\setcounter{footnote}{0}%
\renewcommand{\thefootnote}{\fnsymbol{footnote}}%
\footnote{#1}%
\let\thefootnote=\oldthefootnote%
}

\begin{document}


\title{EFI: A Toolbox for Feature Importance Fusion and Interpretation in Python}

\author{Aayush Kumar\inst{1,*} \and
Jimiama M. Mase\inst{1,*} \and
Divish Rengasamy\inst{1} \and
Benjamin Rothwell\inst{1} \and
Mercedes Torres Torres\inst{2} \and
David A. Winkler\inst{3} \and
Grazziela P. Figueredo\inst{1}}

\authorrunning{A. Kumar et al.}

\institute{The University of Nottingham, UK \and
B-Hive Innovations, UK \and
Monash Institute for Pharmaceutical Sciences, Australia}

\maketitle
\begin{abstract}
\astfootnote{First and second authors have equally contributed to the work.}This paper presents an open-source Python toolbox called Ensemble Feature Importance (EFI) to provide machine learning (ML) researchers, domain experts, and decision makers with robust and accurate feature importance quantification and more reliable mechanistic interpretation of feature importance for prediction problems using fuzzy sets. The toolkit was developed to address uncertainties in feature importance quantification and lack of trustworthy feature importance interpretation due to the diverse availability of machine learning algorithms, feature importance calculation methods, and  dataset dependencies. EFI merges results from multiple machine learning models with different feature importance calculation approaches using data bootstrapping and decision fusion techniques, such as mean, majority voting and fuzzy logic. The main attributes of the EFI toolbox are: (i) automatic optimisation of ML algorithms, (ii) automatic computation of a set of feature importance coefficients from optimised ML algorithms and feature importance calculation techniques, (iii) automatic aggregation of importance coefficients using multiple decision fusion techniques, and (iv) fuzzy membership functions that show the importance of each feature to the prediction task. The key modules and functions of the toolbox are described, and a simple example of their application is presented using the popular Iris dataset.


\end{abstract}

\section{Introduction}
\label{intro}
Machine Learning (ML) systems are providing very useful autonomous and intelligent solutions in diverse science and technology domains. In areas where the decisions from the ML systems ultimately affect human lives (e.g., healthcare, transport, law and security), it is important to understand how and why  the decisions are made for system verification~\cite{arrieta2020explainable}, regulatory compliance~\cite{Reddy2019}, elucidation of ethical concerns, trustworthiness~\cite{GILLE2020100001}, and system diagnostics~\cite{rengasamy2021towards}. A potential solution to the problem of understanding system decisions is model interpretability that aims to understand decisions of complex ML architectures not readily interpretable by design (e.g. neural networks, support vector machines and ensembles of decision trees). A popular approach used in model interpretation is feature importance (FI) analysis, which estimates the contribution of each data feature to the model's output~\cite{rengasamy2021towards}. Several comprehensive open-source libraries already exist for ML automation~\cite{autosklearn,expresso,autoMLtables} and explainability~\cite{alibi,dalex,mlexplainability360,interpretML}. However, the available libraries do not cover ensemble feature importance and fuzzy logic interpretability.

The wide availability of ML algorithms and diversity of FI techniques complicates the selection of ML and FI approaches and the reliability of interpretations. Different ML models may generate different FI values due to variations in their learning algorithms. Similarly, different FI techniques may produce different importances for the same ML algorithms, and different data samples may produce different importances for the same ML algorithms and FI techniques. Additionally, for models in which the response surface is significantly nonlinear (curved in multidimensional space), FI is a local rather than global property. 

To address these uncertainties in the selection and interpretation of models, ensemble feature importance (EFI) methods have been proposed that combine the results from multiple ML models coupled with different feature importance quantifiers to produce more robust, accurate, and interpretable estimates of FI~\cite{rengasamy2021towards,rengasamy2021mechanistic,bobek2021towards,zhai2018development,ruyssinck2014nimefi,huynh2021optimizing}. The latest ensemble methods apply multiple model-agnostic feature importance methods to multiple trained ML algorithms, and aggregate the resulting FI coefficients using crisp~\cite{rengasamy2021towards} or/and fuzzy~\cite{rengasamy2021mechanistic} decision fusion strategies.

The fuzzy ensemble feature importance (FEFI) method extends the crisp ensemble methods by modelling the variance in feature importance generated by the different ML methods, FI techniques, and data spaces using fuzzy sets. FEFI has shown to provide more reliable FI values for high-dimensional datasets with non-linear feature relationships, and noise injected by system dynamics~\cite{rengasamy2021mechanistic}. In addition, FEFI provides better explanations of FI using linguistic terms: `low', `moderate' and `high' importance.

To further enhance the utility of fuzzy approaches to FI, here we present an Ensemble Feature Importance (EFI) toolbox, developed in the Python programming environment (available online\footnote{https://github.com/jimmafeni/EFI-Toolbox}) that implements the crisp and fuzzy ensemble feature importance strategies. It includes automatic data preprocessing, model training and optimisation, ensemble feature importance calculation, and feature importance interpretation. The following sections present the main modules and functions of EFI toolbox as an approach to assist ML researchers, domain experts, and decision makers to obtain reliable interpretations of the importance of features. It is also a diagnostic tool for identifying data subsets with extreme cases of feature importance or with significant variation of feature importance.

\section{EFI Toolbox}
\label{toolkit}

\begin{figure*}[!hb]
\centering
\scalebox{0.65}
{
\includegraphics{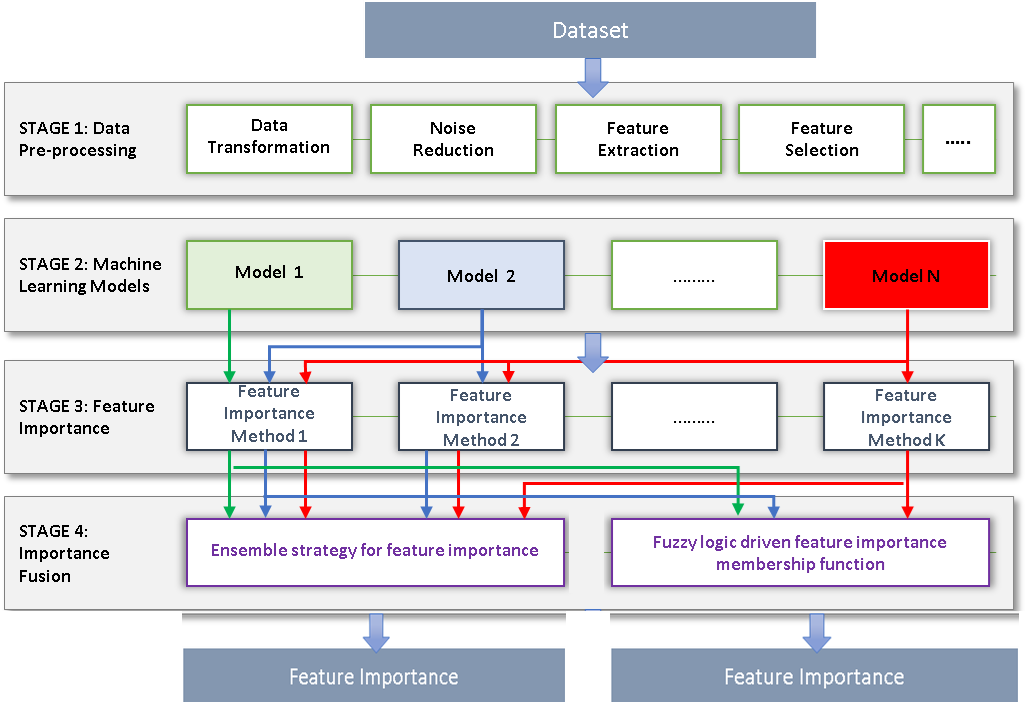}
}
\caption{The four stages of EFI framework. }
\label{fig:fuzzysystem}
\end{figure*}

\subsection{An overview of EFI framework}

Currently, the EFI toolbox is implemented for classification problems based on the EFI framework proposed for crisp~\cite{rengasamy2021towards} and fuzzy ensemble strategies~\cite{rengasamy2021mechanistic}. However, the toolbox can be easily adapted for regression problems. The EFI framework generates a set of FI coefficients using an ensemble of ML models coupled with multiple FI techniques and data bootstrapping. The FI coefficients are aggregated using crisp decision fusion strategies to produce robust and accurate FIs or/and using fuzzy ensemble (FEFI) for a more robust interpretation of FIs. A flowchart of the EFI framework is shown in Fig.~\ref{fig:fuzzysystem} and the four stages of the process are described in the following subsections. 

\subsubsection{Stage 1: Data Preprocessing}
Data are preprocessed to manage missing values, outliers, irrelevant features, and highly correlated features. Next, they are normalised to ensure that all features are of the same magnitude. The preprocessed data are partitioned into $k$ equally sized data samples for model optimisation i.e., $k$-fold cross validation. 

\subsubsection{Stage 2: Machine Learning Model Optimisation}
Here, an ensemble of optimised ML models is generated using cross-validation and hyperparameter tuning (e.g., grid search and random search) on multiple ML algorithms. This is important to ensure that the most accurate models are obtained for the specified problem.

\subsubsection{Stage 3: Feature Importance Coefficients}
Feature importance approaches are applied to the optimised, trained models to compute importance coefficients. For model-agnostic feature importance techniques, validation data subsets are used to produce the coefficients. The importance coefficients obtained for each ML-FI pair are normalised to the same scale to ensure unbiased and consistent representation of importance. The output of this stage is a set of normalised FI coefficients. 

\subsubsection{Stage 4: Importance Fusion}
Here, we consider crisp and fuzzy ensemble strategies as follows:
\begin{itemize}
    \item Crisp ensemble feature importance: The coefficients from the ensemble of ML models using multiple FI techniques are aggregated using crisp decision fusion methods such as mean, median, mode etc. (denoted here as multi-method ensemble feature importance). Note that the aggregated coefficients of individual ML algorithms can also be obtained by filtering coefficients from a particular ML algorithm (denoted here as model-specific ensemble feature importance).
    
    \item Fuzzy ensemble feature importance: FEFI generates membership functions (MFs) from the set of coefficients, labelling the importance of features as `low', `moderate' and `high'. For each ML approach, FEFI assigns the labels low, moderate and high importance to the feature set. Each feature is assigned  low, moderate and high importance relative to the other features in the data for each ML approach and for all ML approaches combined. 
    
\end{itemize}

\begin{figure}
\centering
\scalebox{0.37}
{
\includegraphics{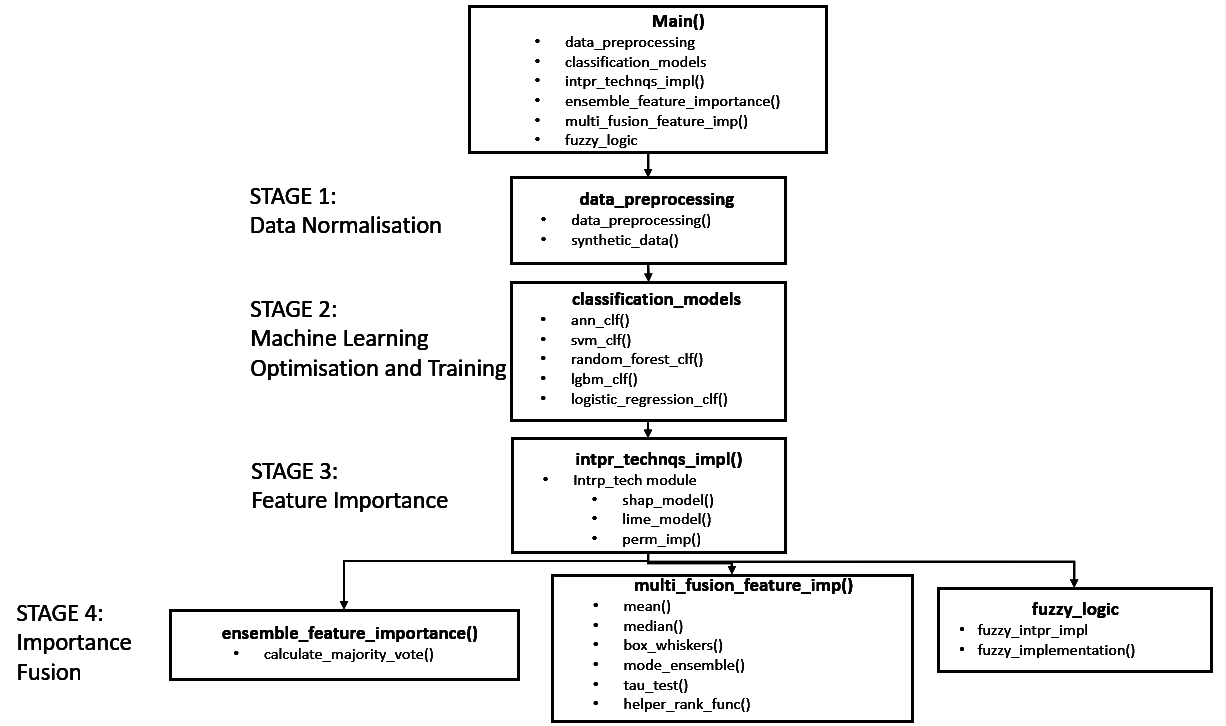}
}
\caption{Layout of EFI toolbox.}
\label{fig:fefitoolkit}
\end{figure}

\subsection{Toolbox Modules}
\label{toolbox features}
A modular programming approach is employed in developing the toolbox for easy documentation, debugging, testing, and extensibility as shown in Fig.~\ref{fig:fefitoolkit}, with a \textit{Main()} module for executing the various stages of EFI framework. The main modules of EFI toolbox are: – 

\subsubsection{Data Preprocessing}
The essential task of automatic preprocessing of the data (i.e. normalise and encode target labels) occurs in the data\_preprocessing module. This module can also generate synthetic data, using synthetic\_data(),  for testing the toolbox . The following is an example of the application of this module, where a feature set \textit{X} is preprocessed and target labels \textit{Y} are encoded:

 \begin{equation*} 
    \centering
    \begin{aligned}
    \color{blue}x,y = data\_preprocessing  & \color{blue}(X,Y,\\
                        & \color{blue}loc\_specified='No')\\
    \end{aligned}
\end{equation*}
    
\noindent A user can also pass the data location and target column name as \textit{X} and \textit{Y} parameters and set \textit{loc\_specified} to `Yes'.

\subsubsection{Model Training and Optimisation}
The \textit{classification\_models} module automatically optimises ML algorithms using grid search hyperparameter tuning and cross validation. The best performing hyperparameters are used to train and return the model. A user can choose which ML algorithm to optimise and train by  passing as parameters the dataset and number of folds, as shown in the command below:

 \begin{equation*} 
    \centering
    \begin{aligned}
    \color{blue}ann = \ ann\_clf  & \color{blue}(X\_train,Y\_train,\\
                        & \color{blue}X\_test,Y\_test,k)\\
    \end{aligned}
\end{equation*}

\noindent The command returns an optimised neural network by performing \textit{k}-fold cross-validation on the training set ($X$\_$train$, $Y$\_$train$). The optimised model is evaluated using ($X$\_$test$, $Y$\_$test$) and its performance (classification accuracy and ROC-AUC curve) is displayed and saved in output files. There exist similar functions in \textit{classification\_models} module for optimising Gradient Boost, Logistic Regressor, Random Forests and Support Vector Machine classifiers.

\subsubsection{Calculate feature importance coefficients}
Model-agnostic FI methods are run using the function \textit{intpr\_technqs\_impl}(). This function takes  as parameters a dataset, a trained ML model, data split size for computing importance coefficients, and the name of the model. It  returns a set of normalised FI coefficients. It partitions the data according to the data split size parameter and uses the partitioned data to calculate SHapley Additive exPlanations (SHAP), Local Interpretable Model-agnostic Explanations (LIME), and Permutation Importance (PI) importance coefficients on the trained model as shown in the command below:

\begin{equation*} 
\centering
\begin{aligned}
\color{blue}SHAP, \ LIME, \ PI = \ & \color{blue}intpr\_technqs\_impl(X,\\
                      & \color{blue}Y, LR\_model, data split size=0.2,\\
                      &\color{blue}'Logistic Regression')
\end{aligned}
\end{equation*}

\noindent The above command returns the SHAP, LIME, and PI importance coefficients of the features in dataset \textit{(X,Y)} when 80\% of the dataset is used to train and optimise a logistic regression model and the remaining 20\% of the dataset is used to calculate FI coefficients.

The module uses functions within \textit{intrp\_tech} module (i.e., shap\_model(), lime\_model(), perm\_imp()) to calculate the importance coefficients for SHAP, LIME, and PI respectively. For example,

\begin{equation*} 
\centering
\begin{aligned}
\color{blue}PI = \ perm\_imp & \color{blue}(X,Y,ANN\_model\\
                      & \color{blue}model\_name) \\
\end{aligned}
\end{equation*}

\noindent returns PI importance coefficients of the features in dataset \textit{(X,Y)} on the trained ML model (i.e ANN\_model). These functions are implemented using Python functions for model-agnostic feature importance quantification:
\begin{itemize}
    \item \textit{permutation\_importance():} This function is found in sklearn.inspection module. It calculates the importance of features in a given dataset by measuring the increase in the prediction error of the trained model after permuting the values of features~\cite{rengasamy2021towards}. 
    
    \item \textit{KernelExplainer():}  KernelExplainer function is found in the shap library for calculating the SHAP values for a set of samples. SHAP determines the average marginal contribution of features to a model's prediction by considering all possible combinations of the features~\cite{rengasamy2021towards}. 
    
     \item \textit{LimeTabularExplainer(.):}  This function is found in lime library for calculating the LIME values of features of randomly selected instances in the dataset. LIME determines the average contribution of features of randomly selected instances in the dataset by generating a new dataset consisting of perturbed samples with corresponding predictions from the original model, and training interpretable models on the new dataset~\cite{ribeiro2016model}. The perturbed samples are weighted according to their proximity from the randomly selected instances. 
    
\end{itemize}

\subsubsection{Model-specific ensemble feature importance}
After calculating the feature importance coefficients, the toolbox provides a function to combine the feature importance coefficients for individual ML algorithms using a majority vote ensemble method i.e. $ensemble$\_$feature$\_$importance$(). This model-specific ensemble method is useful to obtain the importance of features for a specific ML algorithm with higher classification accuracy compared to the other ML algorithms. For example, the command

\begin{equation*} 
\centering
\begin{aligned}
\color{blue}LR\_fi = \ & \color{blue}ensemble\_feature\_importance(SHAP,\\
                        & \color{blue} LIME,PI, model\_name,num\_features)\\
\end{aligned}
\end{equation*}

\noindent combines the importance coefficients obtained by applying \textit{SHAP}, \textit{LIME}, and \textit{PI} techniques on the ML algorithm specified using 'model\_name' parameter. For example, if the $model$\_$name$ is equal to 'Logistic Regression', the command will return the aggregated importance coefficients for the Logistic Regression classifier. The value of the parameter num\_features ranges from 0 to 1  and it determines the number of features to consider in the ranking process of majority vote ensemble~\cite{rengasamy2021towards}.

\subsubsection{Multi-method ensemble feature importance}
The EFI toolbox provides a function, \textit{multi\_fusion\_feature\_imp}(),  for combining the importance coefficients from multiple ML algorithms using multiple FI techniques. This  is suitable for problems where the ML algorithms do not provide required predictive performance and there exist significant variations in the prediction of instances and the quantification of FIs. With such problems, a multi-method ensemble usually achieves more accurate and robust interpretations of FIs. For example, this command:

\begin{equation*} 
\centering
\begin{aligned}
\color{blue}fi\_df = \ & \color{blue}multi\_fusion\_feature\_imp(SHAP,\\
                        & \color{blue} LIME,PI, X,models\_selected)\\
\end{aligned}
\end{equation*}

\noindent combines the importance coefficients obtained by applying \textit{SHAP}, \textit{LIME}, and \textit{PI} techniques on the ML algorithms specified using the $models$\_$selected$ parameter. Therefore, if '$models$\_$selected$' is equal to ['Logistic Regression','Random Forests'], the feature will combine the importance coefficients obtained from both models. 

The following decision fusion methods are employed within this module: Mode, Median, Mean, Box-Whiskers, Tau Test, Majority Vote, RATE-Kendall Tau, RATE-Spearman Rho. For more information regarding these methods, please refer to~\cite{rengasamy2021towards}. The methods are implemented using the following functions within \textit{multi\_fusion\_feature\_imp}():

\begin{itemize}

    \item \textit{majority\_vote\_func}(): Ranks features based on their coefficients produced by each ML model and FI technique. Calculate the mean of the coefficients in the most common ranks for each feature.   
    
    \item \textit{mean}(): Averages importance coefficients obtained by FI techniques for each feature.
    
    \item \textit{median}(): Sorts importance coefficients for each feature and returns the middle coefficients as the final coefficient of features. 

    \item \textit{box\_whiskers}(): Averages coefficients that lie between the lower and upper whiskers of their distribution. The whiskers are obtained by adding the interquartile range to the third quartile (upper whisker) or subtracting the interquartile range from the first quartile (lower whisker) to eliminate potential outliers in the data.
    
    \item \textit{mode\_ensemble}(): Uses a probability density function to get an estimate of the mode of the coefficients for each feature.
    
    \item \textit{tau\_test}(): Uses Thompson Tau test to detect and remove anomalies, and uses the mean to determine the final coefficient of each feature. 
    
    \item \textit{helper\_rank\_func}(): Uses Kendall and Spearman statistical tests to eliminate outliers and employs a majority vote ensemble to determine the final coefficient of each feature. 
    
\end{itemize}

\subsubsection{Fuzzy ensemble feature importance}
The toolbox provides a module called \textit{fusion\_logic} for combining the importance coefficients using fuzzy logic. This  prevents loss of information and provides a more understandable representation of importance as linguistic terms i.e. `low', `moderate' and `high' importance. The main function in this module, $fuzzy$\_$implementation$(), takes as parameters a dataset of FIs obtained using cross-validation and a list of model names to generate their membership functions~\cite{zadeh1975fuzzy}. For example, the command:

\begin{equation*} 
\centering
\begin{aligned}
\color{blue} fuzzy\_implementation(FIs,model\_names)\\
\end{aligned}
\end{equation*}

\noindent combines the importance coefficients (FIs) obtained by applying feature importance techniques on the ML algorithms to produce membership functions determining:

\begin{enumerate}
    \item for each ML approach listed in `model\_names', the range of low, moderate and high importance in the feature set after training;  
    \item for each feature, its low, moderate, and high importance relative to the other features in the data for each ML approach; 
    \item for each feature, its low, moderate, high importance after training and after combining with importances from multiple ML approaches.
\end{enumerate}

\section{Case Study}
\label{examples}

In this section, we will use the popular Iris dataset to illustrate the application of our toolbox for feature importance fusion and interpretation. 

\subsection{The Iris dataset}
\label{iris dataset section}

\begin{figure}[!ht]
\centering
\scalebox{0.65}
{
\includegraphics{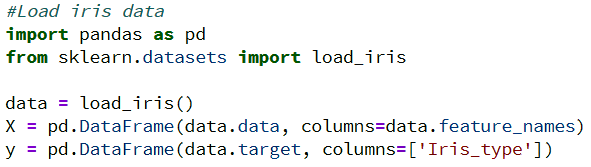}
}
\caption{Code snippet to load Iris data}
\label{fig:irisdataset}
\end{figure}

The Iris dataset is an open-source flower classification dataset that consists of three types of flowers i.e. Setosa, Versicolour, and Virginica. The dataset is made up of 50 samples from each of the three types of iris flowers and for each sample, four features are reported: sepal length, sepal width, petal length and petal width. Fig.~\ref{fig:irisdataset} presents a code snippet to load the dataset in Python.

\subsection{Data pre-processing}
\label{iris dataset}
We normalise the features in the iris dataset and transform all target labels to normalised integer class representations. Fig.~\ref{fig:preprocess} presents a code snippet to pre-process the loaded iris dataset \textit{(X,y)} using our toolbox:

\begin{figure}[!ht]
\centering
\scalebox{0.65}
{
\includegraphics{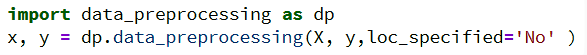}
}
\caption{Code snippet to pre-process Iris data using EFI}
\label{fig:preprocess}
\end{figure}

\subsection{Model optimisation and training}
\label{model optimisation}
\begin{figure}[H]
\centering
\scalebox{0.55}
{
\includegraphics{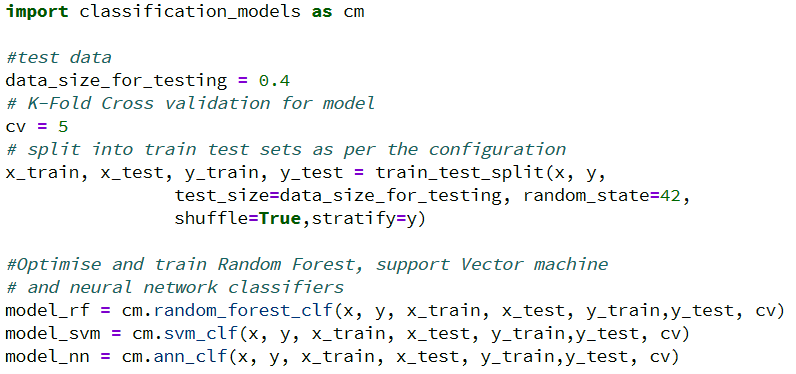}
}
\caption{EFI code to optimise machine learning models}
\label{fig:rfmodel}
\end{figure}

We optimise and train random forest, support vector machine, and neural network models to classify iris flowers as shown in Fig.~\ref{fig:rfmodel}.

\noindent The optimised random forest classifier (model\_rf) has an average accuracy of 90\% and an average f1-score of 88.4\% with hyperparameters: \{`bootstrap': False, `criterion of split': `entropy', `minimum samples at leaf node': 5, `number of trees': 300\}. The optimised support vector machine classifier (model\_svm) produces an average accuracy of 90\% and an average f1-score of 89.5\% with hyperparameters: \{`regularisation coefficient': 1, `kernel': `rbf', `kernel coefficient': 0.1, \}. The optimised neural network model (model\_nn) produces an average accuracy of 91\% and an average f1-score of 83.1\% with hyperparameters: \{`learning rate': 0.001, `batch size': 2, `epochs': 25 \}. 

\subsection{Feature importance coefficients}
\begin{figure}[H]
\centering
\scalebox{0.53}
{
\includegraphics{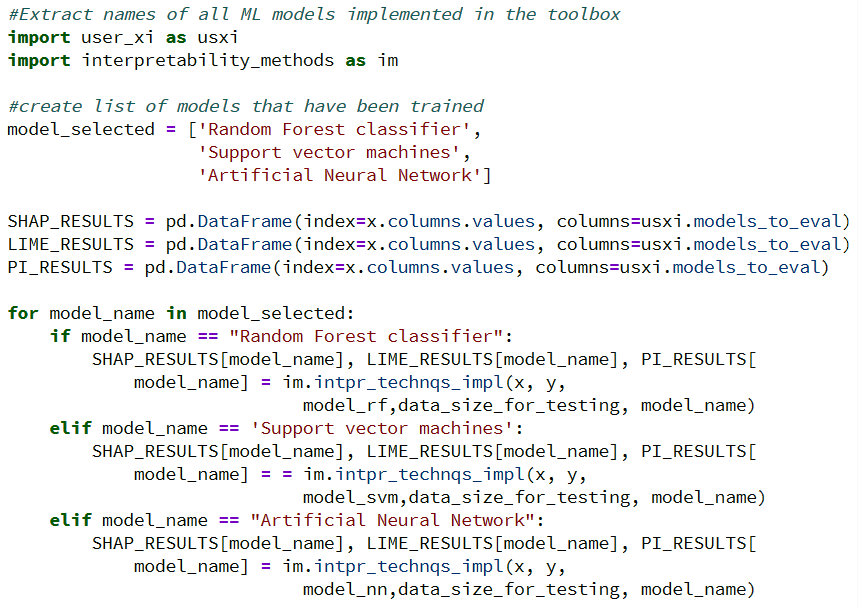}
}
\caption{EFI code to calculate feature importance using SHAP, LIME and PI for multiple machine learning models }
\label{fig:ficalc}
\end{figure}

We calculate  the contribution (importance) of each feature (sepal length, sepal width, petal length and petal width) in classifying the three types of iris flowers by applying the FI techniques on the models trained above. Fig.~\ref{fig:ficalc} illustrates the calculation of FI coefficients of random forest, support vector machines and artificial neural networks. 

\noindent The resulting normalised importance coefficients from each FI technique are shown in Tables~\ref{tab:picoef},~\ref{tab:shapcoef}, and~\ref{tab:limecoef}. The bold coefficients represent the top contributing feature to a model's classification. We observe disagreements in the most important features across ML models and FI techniques. In addition, trusting the importance results of a particular ML algorithm is difficult as all three ML algorithms show similar classification performance, presented in Section~\ref{model optimisation} above.  Hence, the motivation for EFI. \\

\begin{table}[!ht]
  \centering
  \caption{Normalised permutation importance coefficients for Iris features (best contribution in bold).}
  \scalebox{0.9}{
    \begin{tabular}{|c|c|c|c|}
    \hline
         & \multicolumn{1}{c|}{\textbf{Neural Network}} & \multicolumn{1}{c|}{\textbf{Random Forest}} & \multicolumn{1}{c|}{\textbf{Support Vectors}} \\
    \hline
    \textbf{sepal length (cm)} & 0.07 & 0.02 & 0.00 \\
    \hline
    \textbf{sepal width (cm)} & 0.00 & 0.00 & 0.08 \\
    \hline
    \textbf{petal length (cm)} & \textbf{1.00} & 0.45 & 0.91 \\
    \hline
    \textbf{petal width (cm)} & 0.52 & \textbf{1.00} & \textbf{1.00} \\
    \hline
    \end{tabular}}%
  \label{tab:picoef}%
\end{table}%

\begin{table}[!ht]
  \centering
  \caption{Normalised shap coefficients for Iris features (best contribution in bold).}
  \scalebox{0.9}{
    \begin{tabular}{|c|c|c|c|}
    \hline
         & \multicolumn{1}{c|}{\textbf{Neural Network}} & \multicolumn{1}{c|}{\textbf{Random Forest}} & \multicolumn{1}{c|}{\textbf{Support Vectors}} \\
    \hline
    \textbf{sepal length (cm)} & 0.20 & 0.12 & 0.00 \\
    \hline
    \textbf{sepal width (cm)} & 0.00 & 0.00 & 0.01 \\
    \hline
    \textbf{petal length (cm)} & \textbf{1.00} & 0.84 & \textbf{1.00} \\
    \hline
    \textbf{petal width (cm)} & 0.37 & \textbf{1.00} & 0.90 \\
    \hline
    \end{tabular}}%
  \label{tab:shapcoef}%
\end{table}%

\begin{table}[!ht]
  \centering
  \caption{Normalised lime coefficients for Iris features (best contribution in bold).}
  \scalebox{0.9}{
    \begin{tabular}{|c|c|c|c|}
    \hline
         & \multicolumn{1}{c|}{\textbf{Neural Network}} & \multicolumn{1}{c|}{\textbf{Random Forest}} & \multicolumn{1}{c|}{\textbf{Support Vectors}} \\
    \hline
    \textbf{sepal length (cm)} & \textbf{1.00} & 0.11 & 0.58 \\
    \hline
    \textbf{sepal width (cm)} & 0.56 & 0.36 & 0.72 \\
    \hline
    \textbf{petal length (cm)} & 0.46 & \textbf{1.00} & 0.00 \\
    \hline
    \textbf{petal width (cm)} & 0.00 & 0.00 & \textbf{1.00} \\
    \hline
    \end{tabular}}%
  \label{tab:limecoef}%
\end{table}%

\subsection{Model specific ensemble feature importance}
\begin{figure}[H]
\centering
\scalebox{0.6}
{
\includegraphics{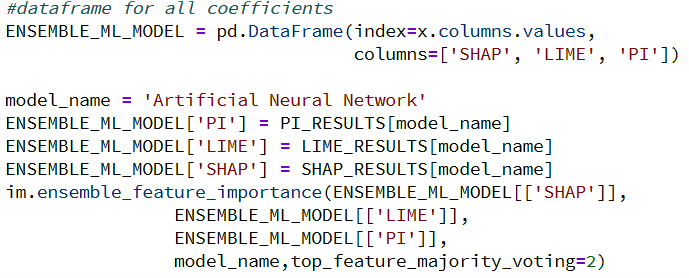}
}
\caption{EFI ensemble of feature importance coefficients for a specific machine learning model}
\label{fig:modelspecific}
\end{figure}

Let's assume we trust the classification performance of the neural network model in predicting iris flowers over the performance of random forest and support vector machine models, we can combine the coefficients obtained using the different FI techniques as shown in Fig.~\ref{fig:modelspecific}. 

\noindent The coefficients obtained from the neural network model are combined using the majority vote ensemble method that ranks features in each FI technique and computes the average of the coefficients in the most common ranks of each feature. This produces the final importance coefficients in Table~\ref{tab:modelspecific}. We can observe that `petal length' is the most importance feature in classifying iris flowers using the neural network model, while `sepal width' is the least important. \\

\begin{table}
\centering
  \caption{Final importance coefficients for Iris features computing using majority vote ensemble method (best contribution in bold).}
    \begin{tabular}{|c|c|}
    \hline
         & \textbf{Artificial Neural Network}  \\
    \hline
    \textbf{sepal length (cm)} & 0.13 \\
    \hline
    \textbf{sepal width (cm)} & 0.00 \\
    \hline
    \textbf{petal length (cm)} & \textbf{1.00} \\
    \hline
    \textbf{petal width (cm)} & 0.45 \\
    \hline
    \end{tabular}%
  \label{tab:modelspecific}%
\end{table}%

\begin{table}
\centering
  \caption{Final importance coefficients for Iris features computing using crisp ensemble methods (best contribution in bold).}
    \scalebox{0.55}{
    \begin{tabular}{|l|c|c|c|c|c|c|c|c|}
    \hline
         & \multicolumn{1}{l|}{\textbf{Box-Whiskers}} & \multicolumn{1}{l|}{\textbf{Majority Vote}} & \multicolumn{1}{l|}{\textbf{Mean}} & \multicolumn{1}{l|}{\textbf{Median}} & \multicolumn{1}{l|}{\textbf{Mode}} & \multicolumn{1}{l|}{\textbf{RATE-Kendall Tau}} & \multicolumn{1}{l|}{\textbf{RATE-Spearman Rho}} & \multicolumn{1}{l|}{\textbf{Tau Test}} \\
    \hline
    \textbf{petal length (cm)} & \textbf{0.74} & \textbf{0.74} & \textbf{0.74} & \textbf{0.91} & \textbf{0.74} & \textbf{1.00} & \textbf{1.00} & \textbf{0.83} \\
    \hline
    \textbf{petal width (cm)} & 0.64 & 0.72 & 0.64 & 0.90 & 0.64 & 0.37 & 0.37 & 0.64 \\
    \hline
    \textbf{sepal length (cm)} & 0.07 & 0.10 & 0.23 & 0.11 & 0.23 & 0.20 & 0.20 & 0.05 \\
    \hline
    \textbf{sepal width (cm)} & 0.19 & 0.18 & 0.19 & 0.01 & 0.19 & 0.00 & 0.00 & 0.00 \\
    \hline
    \end{tabular}}
  \label{tab:multimodel}%
\end{table}%

\subsection{Multi-method ensemble feature importance}
To exploit the advantages of ensemble learning to obtain more robust and accurate importance coefficients for the features in iris dataset, EFI toolbox's multi-ensemble FI method can be used as shown in Fig.~\ref{fig:multimodel}:  

\begin{figure}[H]
\centering
\scalebox{0.65}
{
\includegraphics{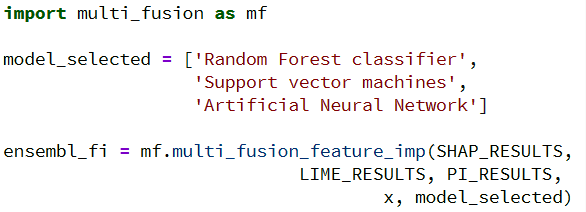}
}
\caption{EFI ensemble of feature importance coefficients of multiple machine learning models}
\label{fig:multimodel}
\end{figure}

\noindent The above code produces the aggregated coefficients in Table~\ref{tab:multimodel} using crisp decision fusion methods. We can observe that `Petal Length' is the most importance feature for all methods. However, the coefficients simply specify the most important feature or importance rank of features. They do not provide an interpretation of how important the features are. \\

\subsection{Fuzzy ensemble feature importance}
An understanding of the importance of each feature to the classification of iris flowers is provided by EFI toolbox's fuzzy ensemble FI method. This method is also important in calculating the importance coefficients for high dimensional datasets, non-linear relationships, and the presence of noise because it explores the data space more thoroughly and uses fuzzy logic to capture uncertainties in coefficients. 

The fuzzy ensemble FI method consist of 2 steps:

\begin{itemize}
    \item Compute FI coefficients for different partitions of the data.
    
    \item Generate membership functions from the coefficients to explain the range of importance for coefficients in each ML algorithm, the importance of each feature relative to the other features in the dataset and compute the aggregated coefficients of features.
    
\end{itemize}

\begin{figure*}[!ht]
\centering
\scalebox{0.35}
{
\includegraphics{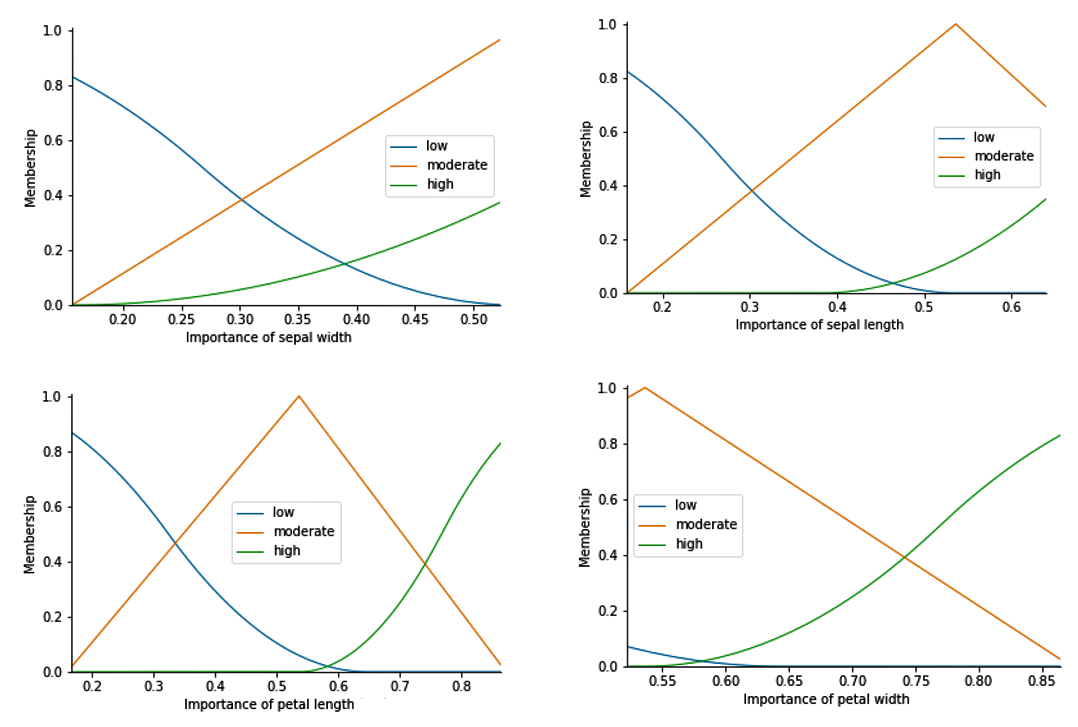}
}
\caption{Membership functions of features generated from Iris flower dataset showing the interpretations of feature importance and their levels of uncertainty}
\label{fig:fuzzyimportance}
\end{figure*}

\noindent The fuzzy ensemble method provides interpretations for the importance of iris flower features as well as their levels of uncertainty as shown in Fig.~\ref{fig:fuzzyimportance}. We observe that both 'petal width' and 'petal length' have high likelihood of `moderate' and `high' importance. However, 'petal width' shows less uncertainty in importance compared to `petal length' i.e., importance coefficients range from 0.50 to 0.85 for 'petal width' and 0.15 to 0.85 for 'petal length'. `sepal width' and `sepal length' have a high likelihood of `low' and `moderate' importances, with importance coefficients ranging from 0.1 to 0.55 and 0.1 to 0.65 respectively. With these membership functions, users can diagnose which data subsets or time steps produce the extreme cases of feature importance.

It is important to note that the code snippets used to demonstrate the application of EFI toolbox on the Iris dataset are extracted from different modules in the EFI toolbox because the implementation of EFI follows a modular software engineering approach. In addition, the toolbox can be easily extended to include other ML algorithms, model specific FI techniques and regression problems. Lastly, the toolbox outputs detailed reports of model optimisation and evaluation, and multiple graphs and plots to provide a clear understanding of computation and fusion of FIs by the different FI techniques and ensemble strategies. 


\section{Conclusion}
\label{conclusion}

This paper has presented a novel, extensible open-source toolbox in Python programming language called EFI (Ensemble Feature Importance) that aggregates feature importance coefficients from multiple models coupled and different feature importance techniques using bootstrapping and decision fusion methods. We have described the major modules and functions of the toolbox and provided a step-by-step example using the popular Iris dataset. The outputs of our toolbox (i.e. feature importance coefficient tables, plots, graphs and membership functions) substantially improve the understanding and interpretation of the importance of features to prediction tasks. Due to the complexity and multitude of steps and methods to: (i) optimise multiple ML algorithms, (ii) calculate and visualise the importance of features using various feature importance techniques, (iii) aggregate the importance coefficients using multiple ensemble methods, and (iv) create fuzzy logic systems for capturing uncertainties and interpreting importance, it is important to have tools to make these tasks as simple and robust as possible. As an open-source toolbox, we plan to extend the toolbox to deal with regression tasks in the future and encourage  other researchers to contribute to its growth, such as by improving the structure of the toolbox, implementing additional state-of-the-art ML algorithms and implementing model specific feature importance techniques.

\bibliographystyle{splncs04}
\bibliography{LODrevised.bib}

\begin{thebibliography}{10}
\providecommand{\url}[1]{\texttt{#1}}
\providecommand{\urlprefix}{URL }
\providecommand{\doi}[1]{https://doi.org/#1}

\bibitem{arrieta2020explainable}
Arrieta, A.B., D{\'\i}az-Rodr{\'\i}guez, N., Del~Ser, J., Bennetot, A., Tabik,
  S., Barbado, A., Garc{\'\i}a, S., Gil-L{\'o}pez, S., Molina, D., Benjamins,
  R., et~al.: Explainable artificial intelligence (xai): Concepts, taxonomies,
  opportunities and challenges toward responsible ai. Information Fusion
  \textbf{58},  82--115 (2020)

\bibitem{mlexplainability360}
Arya, V., Bellamy, R.K.E., Chen, P.Y., Dhurandhar, A., Hind, M., Hoffman, S.C.,
  Houde, S., Liao, Q.V., Luss, R., Mojsilovi\'c, A., Mourad, S., Pedemonte, P.,
  Raghavendra, R., Richards, J., Sattigeri, P., Shanmugam, K., Singh, M.,
  Varshney, K.R., Wei, D., Zhang, Y.: One explanation does not fit all: A
  toolkit and taxonomy of ai explainability techniques (september 2019),
  \url{https://arxiv.org/abs/1909.03012}

\bibitem{dalex}
Baniecki, H., Kretowicz, W., Piatyszek, P., Wisniewski, J., Biecek, P.: dalex:
  Responsible machine learning with interactive explainability and fairness in
  python. Journal of Machine Learning Research  \textbf{22}(214), ~1--7 (2021),
  \url{http://jmlr.org/papers/v22/20-1473.html}

\bibitem{bobek2021towards}
Bobek, S., Ba{\l}aga, P., Nalepa, G.J.: Towards model-agnostic ensemble
  explanations. In: International Conference on Computational Science. pp.
  39--51. Springer (2021)

\bibitem{autosklearn}
Feurer, M., Eggensperger, K., Falkner, S., Lindauer, M., Hutter, F.:
  Auto-sklearn 2.0: Hands-free automl via meta-learning. arXiv:2007.04074
  [cs.LG]  (2020)

\bibitem{GILLE2020100001}
Gille, F., Jobin, A., Ienca, M.: What we talk about when we talk about trust:
  Theory of trust for {AI} in healthcare. Intelligence-Based Medicine
  \textbf{1-2},  100001 (2020).
  \doi{https://doi.org/10.1016/j.ibmed.2020.100001},
  \url{https://www.sciencedirect.com/science/article/pii/S2666521220300016}

\bibitem{autoMLtables}
Google: Auto ml tables. https://cloud.google.com/automl-tables/docs  (Last
  accessed: June 2022)

\bibitem{huynh2021optimizing}
Huynh-Thu, V.A., Geurts, P.: Optimizing model-agnostic random subspace
  ensembles. arXiv preprint arXiv:2109.03099  (2021)

\bibitem{alibi}
Klaise, J., Looveren, A.V., Vacanti, G., Coca, A.: Alibi explain: Algorithms
  for explaining machine learning models. Journal of Machine Learning Research
  \textbf{22}(181), ~1--7 (2021), \url{http://jmlr.org/papers/v22/21-0017.html}

\bibitem{interpretML}
Nori, H., Jenkins, S., Koch, P., Caruana, R.: Interpretml: A unified framework
  for machine learning interpretability. arXiv preprint arXiv:1909.09223
  (2019)

\bibitem{Reddy2019}
Reddy, S., Allan, S., Coghlan, S., Cooper, P.: A governance model for the
  application of {AI} in health care. Journal of the American Medical
  Informatics Association : {JAMIA}  \textbf{27}(3),  491--497 (2019)

\bibitem{rengasamy2021mechanistic}
Rengasamy, D., Mase, J.M., Torres, M.T., Rothwell, B., Winkler, D.A.,
  Figueredo, G.P.: Mechanistic interpretation of machine learning inference: A
  fuzzy feature importance fusion approach. arXiv preprint arXiv:2110.11713
  (2021)

\bibitem{rengasamy2021towards}
Rengasamy, D., Rothwell, B.C., Figueredo, G.P.: Towards a more reliable
  interpretation of machine learning outputs for safety-critical systems using
  feature importance fusion. Applied Sciences  \textbf{11}(24),  11854 (2021)

\bibitem{ribeiro2016model}
Ribeiro, M.T., Singh, S., Guestrin, C.: Model-agnostic interpretability of
  machine learning. arXiv preprint arXiv:1606.05386  (2016)

\bibitem{ruyssinck2014nimefi}
Ruyssinck, J., Huynh-Thu, V.A., Geurts, P., Dhaene, T., Demeester, P., Saeys,
  Y.: Nimefi: gene regulatory network inference using multiple ensemble feature
  importance algorithms. PLoS One  \textbf{9}(3),  e92709 (2014)

\bibitem{expresso}
Wang, Y., Chen, T., Xu, H., Ding, S., Lv, H., Shao, Y., Peng, N., Xie, L.,
  Watanabe, S., Khudanpur, S.: Espresso: A fast end-to-end neural speech
  recognition toolkit. In: 2019 IEEE Automatic Speech Recognition and
  Understanding Workshop (ASRU) (2019)

\bibitem{zadeh1975fuzzy}
Zadeh, L.A.: Fuzzy logic and approximate reasoning. Synthese  \textbf{30}(3),
  407--428 (1975)

\bibitem{zhai2018development}
Zhai, B., Chen, J.: Development of a stacked ensemble model for forecasting and
  analyzing daily average pm2. 5 concentrations in beijing, china. Science of
  The Total Environment  \textbf{635},  644--658 (2018)

\end{thebibliography}

\end{document}